\pdfoutput=1
\documentclass[11pt]{article}
\usepackage{emnlp2021}

\usepackage{times}
\usepackage{latexsym}
\usepackage{graphicx}
\usepackage{booktabs}
\usepackage{comment}
\usepackage{multirow}
\usepackage{enumitem}
\usepackage{xcolor}
\usepackage{amsmath}
\usepackage[T2A,T1]{fontenc}
\usepackage[utf8]{inputenc}

\usepackage{CJKutf8}

\usepackage{microtype}

\title{Preventing Author Profiling through Zero-Shot Multilingual Back-Translation}
\author{David Ifeoluwa Adelani, Miaoran Zhang, Xiaoyu Shen, Ali Davody, \\ \textbf{Thomas Kleinbauer, and Dietrich Klakow}\\
Spoken Language Systems Group, Saarland Informatics Campus, Saarland University, Germany\\
\texttt{\{didelani,mzhang,xshen,adavody\}@lsv.uni-saarland.de} \\
\texttt{\{thomas.kleinbauer,dietrich.klakow\}@lsv.uni-saarland.de}
}
\begin{document}
\maketitle
\begin{abstract}
Documents as short as a single sentence may inadvertently reveal
sensitive information about their authors, including e.g. their gender
or ethnicity. Style transfer is an effective way of transforming texts
in order to remove any information that enables author
profiling. However, for a number of current state-of-the-art
approaches the improved privacy is accompanied by an undesirable drop
in the down-stream utility of the transformed data.

In this paper, we propose a simple, zero-shot way to effectively lower the risk of author
profiling through multilingual back-translation using off-the-shelf translation models. We
compare our models with five representative text style transfer models
on three datasets across different domains. Results from both an
automatic and a human evaluation show that our approach achieves the
best overall performance while requiring no training data. We are able to lower the adversarial prediction of gender and race by up to $22\%$ while retaining $95\%$ of the original utility on downstream tasks. 
\end{abstract}

% Where is the translation system come from, using off-the-shelf

\begin{comment}
% Xiaoyu 
Recent rapid development of large-scaled pre-trained language models has raised a lot of concerns about user privacy leakage. Private information like gender, race or political preferences are captured in the text and can be easily inferred with neural network models. Most approaches address this problem via text style transfer, where text are transformed to change the original user attributes. In this work, we propose a simple yet effective way to perform such transformation in a zero-shot way though multilingual back-translation. On three datasets across different domains, we compare our model against 5 representative text style transfer models with both automatic and human evaluation. We show our approach achieves the best overall performance, albeit requiring zero training data. It is able to significantly lower down the privacy leakage while maintaining the content and downstream utility to a large extent. Text style transfer models, on the contrary, strongly sacrificed the text quality, making it hard to apply in real-life applications.
\end{comment}

\section{Introduction}
\label{sec:intro}

Data collections of natural language utterances bear the risk of disclosing sensitive information about the recorded participants, including their gender, race, or political preferences. Unlike explicit mentions of private information, like %\eg 
a user's name or location \cite{Tang2004PreservingPI,AdelaniDKK20}, such user traits are often encoded rather subtly in a user's speaking or writing style. Nevertheless, they can be predicted with high accuracy by deep learning-based classifiers even when they are not obvious to humans \citep{elazar-goldberg-2018-adversarial}, enabling third-parties with access to the data sets to profile users without their knowledge.

A common method to alleviate this problem is the application of an intermediate transformation step to remove sensitive information via text style transfer. While a number of different style transfer techniques exist \cite{Shen2017_cae,Fu2018StyleTI,madaan-etal-2020-politeness}, they require large amounts of text data labeled with user trait information to perform well. Additional annotations need to be provided for every new user trait that the model is expected to handle, multiplying the associated costs and effort. Furthermore, the impact that such transformations can have on the utility of the resulting data is often overlooked. Conversely, we argue that the privacy-utility dichotomy should be at the heart of all research on this topic because it is fairly easy to consider one of the two but difficult to improve both at the same time. 
\begin{comment}
Thomsa: Two popular approaches that fall into this category are \emph{style transfer}~\cite{Shen2017_cae,He2020A} and \emph{back-translation}. For style transfer, a model is trained to explicitly rewrite the input text to change the user traits that need to be protected. Being a supervised technique, however, this requires large amounts of text data labeled with user trait information to perform well. Moreover, additional annotations need to be provided for every new user trait that the model is expected to handle, multiplying the associated costs and effort.

Back-translation is an alternative approach without such prerequisites. Sensitive user traits can be significantly obfuscated when translated to another language and back \citep{rabinovich-etal-2017-personalized,prabhumoye-etal-2018-style} since many concepts cannot easily be mapped across languages. For example, in languages such Japanese and Korean the speaker's gender can be inferred from the choice of certain pronouns. When back-translating them via an intermediate language that does not make such differences, such as %\eg 
English, these gender indicators will be largely obfuscated. Unfortunately, existing work using back-translation does not achieve the quality of style-transfer methods.

In this paper, we explore a simple but effective zero-shot text transformation method based on multilingual back-translation (BT). Like previous monolingual approaches, it requires no training data.
\end{comment}

In this paper, we explore a simple yet effective zero-shot text transformation method based on multilingual back-translation. Back-translation (BT) is an alternative approach without the prerequisites of labeled training data. Sensitive user traits can be significantly obfuscated when translated to another language and back \citep{rabinovich-etal-2017-personalized,prabhumoye-etal-2018-style} since many concepts cannot easily be mapped across languages. For example, in languages such as Japanese and Korean the speaker's gender can be inferred from the choice of certain pronouns. When back-translating them via an intermediate language that does not make such differences, such as %\eg 
English, these gender indicators will be largely obfuscated. 
%Thomas: Unfortunately, existing work using back-translation does not achieve the quality of style-transfer methods. 

 Results from extensive experiments show that our simple zero-shot text transformer has comparable or even better performance than popular style transfer methods, considering both the privacy and utility of the transformed texts.
In summary, we make the following contributions: 
\begin{enumerate}[leftmargin=0.4cm]
    \item We propose using multilingual back-translation for hiding users traits. We experiment with using 6 high-resourced languages: German, Spanish, French, Japanese, Russian, and Chinese as the pivot language. This provides more opportunities to pick a language that can hide sensitive information represented in the original language. Our approach is zero-shot without the need for additional data to train style transfer models.  
    %\vspace{-2mm}
    \item We show that our simple approach is competitive with style transfer models using automatic metrics, and better performance using human evaluation in terms of content preservation and fluency. 
    %\vspace{-7mm}
    \item We perform a comprehensive evaluation on three datasets with popular style transfer methods. These methods have been well studied in the style transfer community, but they have never been evaluated for both privacy %protection
    and utility preservation in downstream tasks. 
   
    %\item We introduce  evaluation of utility on transformed texts to measure the extent they perform on down-stream NLP task. 
\end{enumerate}

\section{Related Work}
\label{sec:related_work}
%%%%%%%%%%%%%% TODO:  re-write %%%%%%%%%%%
% David
%\paragraph{Attribute Obfuscation} %is a very important research topic in the study of privacy, bias and fairness of deep learning models. 
Attribute information such as gender, age, or race are being captured in the deep learning models. %Some approaches remove these attributes from text representations with adversarial training, differential privacy or both ~\cite{elazar-goldberg-2018-adversarial,coavoux-etal-2018-privacy,ghazaleh_privacy_rep} while others attempt to remove it from the original texts by re-writing the content while preserving the meaning. 
Traditional approaches prevented this information leakage via lexical substitution of sensitive words~\cite{reddy-knight-2016-obfuscating}. In recent years, many text style transfer techniques have been proposed to control certain attributes of generated text (e.g., formality or politeness) while preserving the content. A common paradigm is to disentangle the content and style in the latent space \cite{Shen2017_cae, john-etal-2019-disentangled, cheng-etal-2020-improving}. Another stream of work treats text style transfer as an analogy of unsupervised machine translation \cite{zhang2018style,lample2018multipleattribute,zhao2019unsupervised,He2020A} to rephrase a sentence while reducing its stylistic properties \cite{prabhumoye-etal-2018-style}. Beyond the end-to-end training methods, the prototype-based text editing approach also attracts lot of attention \cite{li-etal-2018-delete,sudhakar-etal-2019-transforming,madaan-etal-2020-politeness}, in which attribute markers of input sentences are deleted and then replaced by target attribute markers. These techniques have been well studied in the text style transfer community, but have never been evaluated for both privacy and utility preservation in downstream tasks.
%One important evaluation task of style transfer models is ``transfer style / attribute strength'' that measures the success of transforming text of one attribute to another. 
\citet{shetty2018} and \citet{xu-etal-2019-privacy} make use of adversarial training and evaluate on authorship obfuscation. However, they did not include most recent style transfer methods and predictors based on pretrained language models. %based on the idea that machine translation (MT) systems often obfuscate attribute information in texts~\cite{rabinovich-etal-2017-personalized}. %\citet{shetty2018} directly applied adversarial training to authorship obfuscation by combining semantic consistency loss, language model loss, and style transfer loss to protect private attributes while ensuring better content preservation. %
%proposes an authorship obfuscation that uses text style transfer to 
% Miaoran
%\paragraph{Text Style Transfer}
 
%More details of style transfer methods are presented in \cite{jin2021deep}.

%- Style transfer approaches: latent representation disentanglement~\cite{Shen2017_cae,prabhumoye-etal-2018-style,shetty2018, Fu2018StyleTI}, online back-translation objective~\cite{lample2018multipleattribute,He2020A}, and delete, retrieve and generate approach~\cite{li-etal-2018-delete,sudhakar-etal-2019-transforming,madaan-etal-2020-politeness}. More details of style transfer methods are in \cite{jin2021deep}. 

\section{Multilingual Back-Translation}
\label{sec:multi_bt}

\paragraph{Problem Scenario} 
In understanding human behaviors and intents, many machine learning applications need to infer important information from users inputs like sentiment, intent, and dialogue act but there is a need to preserve user privacy. 
We consider a scenario where an adversary attempt to predict demographic attributes of user utterances using a pre-trained attribute classification model. We assume that the adversary already has a pre-trained attribute classification model based on publicly available data. Our goal is to transform the original user input text $X$ to $X'$ such that $X'$ (1) prevents the accurate prediction of user attributes, (2) maintains the utility of downstream NLP tasks, (3) maintains the content of $X$ %$T$ 
 and (4) is a fluent text itself.
\begin{table}[t]
 \footnotesize
  %\vspace{-3mm}
  
 \begin{center}
 \scalebox{0.99}{
  \begin{tabular}{cll}
    \toprule
    \textbf{Pivot Language}  & \textbf{Translated} & \textbf{Back-translated}\\ 
    \midrule
    %Original & thank you \textcolor{red}{papi} \\
%\midrule
    DE & Danke Papi & Thank you \textcolor{blue}{daddy} \\
    FR & merci papi & thank you \textcolor{red}{papi} \\
    %BT (RU) & спасибо папи & thanks \textcolor{red}{papi} \\
    ZH & \begin{CJK}{UTF8}{gbsn}谢谢你爸爸\end{CJK} & Thank you \textcolor{blue}{dad}
\\

%\midrule
%    CAE &  this place is definitely very good.
%\\
%    DLS &  i was n't very impressed with this place.\\
%    Tag\&Gen&  this hotel seems to be gorgeous run .
% \\

    \bottomrule
     %\vspace{-7mm}
  \end{tabular}
  }
  
  \caption{\small Multilingual back-Translation of ``thank u \textcolor{red}{papi}'' using DE, FR, ZH as pivot languages. User traits can be obfuscated by choosing the proper pivot language.}
  \label{tab:approach}
  \end{center}
\end{table}

%\begin{figure}[ht]
%     \centering
%     \vspace{-7mm}
%     \includegraphics[width=0.9\linewidth]{figures/back_translation_diagram.pdf}
%     \vspace{-2mm}
%     \caption{Multilingual back-Translation approach for privacy transformation}
%     \label{fig:approach}
% \end{figure}
% \vspace{-3mm}
 In this paper, we explore a simple, zero-shot text transformation method through multilingual back-translation. Our assumption is that, as also supported in previous research~\cite{rabinovich-etal-2017-personalized,prabhumoye-etal-2018-style}, text styles can be significantly obfuscated when being translated to another language (pivot language) then translated back. 
 One example is shown in ~\autoref{tab:approach}. The word ``papi'' is normally used among Latino Americans which exposes their race. When translating them to languages like Chinese then translating back, it becomes the standard form of ``dad'' and thereby protects the user privacy.
%In this paper, we consider multilingual back-translation for the task of zero-shot privacy aware text transformation of user attributes. Previous approaches have combined back-translation with adversarial training to learn attribute-free latent representation~\cite{prabhumoye-etal-2018-style} or prevent risks of attack at the back-translation phase~\cite{xu-etal-2019-privacy}. However, the target language used for back-translation is the \textit{French} language. %One possible reason is that in the early period of machine translation (MT), \textit{English}-\textit{French} have often been used to train MT models. 
%Nowadays, there are enough data for several high-resource languages (HRLs)%\footnote{\url{https://opus.nlpl.eu/}} 
%that can be used for this purpose.
Specifically, we define our text transformation function as:
\begin{equation*}
    X'= T_{L\rightarrow en}(T_{en\rightarrow L}(X))
\end{equation*}
%Although, previous research have considered French as the pivot language~\cite{prabhumoye-etal-2018-style,xu-etal-2019-privacy}, 
where $L$ is the pivot language and $T$ is a translation model. We make use of mBART50\footnote{\url{https://huggingface.co/facebook/mbart-large-50-many-to-many-mmt}} --- an off-the-shelf machine translation model implemented by HuggingFace~\cite{wolf-etal-2020-transformers}. We consider 6 high-resourced languages as the pivot, so as to ensure a decent quality of machine translation models. The languages chosen are German (DE), Spanish (ES), French (FR), Japanese (JA), Russian (RU), and Chinese (ZH) based on the large amount of resources they have on OPUS~\cite{tiedemann-2012-parallel} and Common Crawl corpora~\footnote{\url{https://commoncrawl.org/}}. %Our approach is very simple, we translate a sentence \textit{from English to one of the HRLs, and translate back to English language}.
%We explored several multilingual pre-trained MT models: OPUS-MT~\cite{tiedemann-thottingal-2020-opus}, M2M-100~\cite{fan2020beyond}, and mBART50~\cite{Tang2020MultilingualTW}. OPUS-MT supports over 1,200 translation directions for 150 languages but the ``EN-JA'' direction was missing. M2M-100 and mBART50 have similar performance. For this experiment, we make use of mBART50\footnote{\url{https://huggingface.co/facebook/mbart-large-50-many-to-many-mmt}} for the back-translation experiments using HuggingFace~\cite{Wolf2019HuggingFacesTS}. %Each of the languages has over 10M parallel sentences used for training the multilingual MT model. 

\section{Experiments and Results}
\label{sec:experiments}

\subsection{Datasets}
\label{sec:data}

In this paper, we conduct experiments on three datasets: DIAL \cite{blodgett-etal-2016-demographic}, VerbMobil \cite{weilhammer-etal-2002-multi} and Yelp \cite{reddy-knight-2016-obfuscating, Shen2017_cae}. These datasets comprise of a variety of domains with either race or gender as the sensitive attribute and they also have annotations for dialog acts and sentiment classification that we use to test the utility of downstream NLP tasks. %using annotated with additional sentiment or dialog act information provided in the datasets. 
For \textbf{Yelp}, we find two datasets previously used in the style transfer literature, one for gender (YelpGender)~\cite{reddy-knight-2016-obfuscating} and the other for sentiment (YelpSentiment)~\cite{Shen2017_cae}. The texts are from the same source but each review do not have both gender and sentiment labels. By automatically comparing each review in the test set of YelpGender with the YelpSentiment Dev and Test sets, we created a new \textbf{Dev} set and \textbf{Test} set with 4K reviews, each with both gender and sentiment information. This can be used for future research to evaluate the utility of Yelp Gender dataset. The dataset is available on Github\footnote{\url{https://github.com/uds-lsv/author-profiling-prevention-BT}}.

\autoref{tab:datasets} shows the data splits for three datasets: \textbf{Attribute Train}, training set for attribute classification; \textbf{Utility Train}, training set for a downstream NLP task; 
\textbf{Style Train}, training set for style transfer, \textbf{Dev}, the development set, and the \textbf{Test} set. The detailed data description is in \autoref{sec:appendix_data}. %The training data for all the splits are disjoint except for \textit{Style Train} in the Twitter and VerbMobil corpora.  
\begin{table}[t]
 \footnotesize
  \vspace{-1mm}
 \begin{center}
 \scalebox{0.93}{
  \begin{tabular}{@{}l@{\hspace{2pt}}rrrp{3mm}r@{}}
    \toprule
    & \textbf{Attribute} &\textbf{Utility} &\textbf{Style} &  &  \\
    \textbf{Dataset}  & \textbf{Train} & \textbf{Train} & \textbf{Train} & \textbf{Dev} &  \textbf{Test} \\ 
    \midrule
    DIAL (race) & 80K & 100K & 100K  & 4K & 4K \\
    %VerbM (gender) & 5442 & 4977 & 5442 & & 1096 \\
    VerbMobil (gender) & 5K & 4977 & 5K & 442 & 1096 \\
    Yelp (gender)& 2.6M & 373K & 200K & 4K & 4K  \\
   % YelpS (gender) & 2.6M & 373K & 5K & 500 & 4K \\
    
    \bottomrule
  \end{tabular}
  }
  
  \caption{Data splits for DIAL, VerbMobil, and Yelp. The utility task for Yelp and DIAL is sentiment classification while for VerbMobil is dialog act classification. }
  \label{tab:datasets}
  \end{center}
  \vspace{-3mm}
\end{table}

\subsection{Experimental Setup}
We train five popular style transfer methods: %(1) Adv~\cite{goodfellow_gan} (2) SMDSP~\cite{xu-etal-2019-privacy}
1) CAE~\cite{Shen2017_cae}, (2) BST~\cite{prabhumoye-etal-2018-style}, (3) UNMT~\cite{lample2018multipleattribute}, (4) DLS~\cite{He2020A}, and (5) Tag\&Gen~\cite{madaan-etal-2020-politeness}. CAE and BST are based on latent representation disentanglement through adversarial training. UNMT and DLS make use of the unsupervised machine translation objective. Tag\&Gen is based on prototype-based text editing using frequency ratios method to tag appropriate attribute markers, and generate replacements with a transformer language model. %SMDSP comes from the fairness community that is used to quantify the discrepancy of demographic parities by measuring maximal deviation between subgroup predictions (MDSP)~\cite{calmon_msdp}. 
We compare the performance of the style transfer models with multilingual BT models based on mBART50. In addition, we compare with reported results in ~\citet{xu-etal-2019-privacy} on the DIAL dataset. For the attribute, sentiment and dialog act classification, we fine-tune a BERT-base~\cite{devlin-etal-2019-bert} model end-to-end. % as described in \autoref{sec:multi_bt}.

\begin{table}[t]
 \footnotesize
  \vspace{-3mm}
 \begin{center}
 \scalebox{0.9}{
  \begin{tabular}{p{12mm}rrrrr}
    \toprule
    & \textbf{Attr.} &\textbf{Util.}  & & & \textbf{} \\
    \textbf{Method} & \textbf{F1$\downarrow$} &\textbf{F1$\uparrow$}  & \textbf{METEOR$\uparrow$} & \textbf{GAR$\uparrow$}  & $\bf{P_{Mean}}\uparrow$ \\
    \midrule
    Original Test set & 88.79 &  75.13 &100  &  48.40 & 58.69\\
    \midrule
    BT (DE) & 81.37  & \textbf{73.84} & \textbf{47.47} & 51.83 & 47.94 \\
    BT (ES) & 69.44 & 70.33 & 32.76 & 63.50 & 49.29 \\
    BT (FR) & 77.72 & 72.60 & 41.78 & 54.88 & 47.89  \\
    BT (JA) & 73.77 & 72.00 & 34.63 & 62.22 & 48.77  \\
    BT (RU) & 78.81 & 73.00 & 42.98 & 50.38 & 46.89  \\
    BT (ZH) & 66.65 & 71.68 & 27.61 & \textbf{80.95} & \textbf{53.40} \\
    \midrule
    %Adv ($\alpha=1$)~\cite{xu-etal-2019-privacy} & \_ / 74.08 & \_ / 70.15 & 18.07 & 27.73 \\
    Adv & 65.75 & 65.70 & 17.03 & \_ & \_ \\
    SMDSP & 74.85 & 69.88 & 28.15 & \_ & \_  \\
    %SMDSP ($\alpha=10$)~\cite{xu-etal-2019-privacy} & \_ / 74.08 & \_ / 70.60 & 26.99 & 33.40 \\
    
    \midrule
    CAE & 35.37& 61.63& 12.84 &  22.08 & 40.30  \\
    BST & \textbf{13.99} &  54.16 & 5.03 & 10.60 & 38.95   \\
    UNMT & 18.11& 64.68 & 19.95 & 43.87 & 52.60 \\
    DLS& 28.13 & 66.18 & 25.04 & 30.28 & 48.34 \\
    Tag\&Gen & 44.34& 69.74& 42.30 & 23.18 & 47.72  \\
    \bottomrule
     \vspace{-3mm}
  \end{tabular}
  }
  \caption{Evaluation on DIAL dataset. Adv and SMDSP result are from \cite{xu-etal-2019-privacy}}
  \label{tab:twitter_result}
  \end{center}
\end{table}

\begin{table*}[t]
 \footnotesize
  \vspace{-3mm}
 \begin{center}
   %\resizebox{\textwidth}{!}{
   \scalebox{0.90}{
  \begin{tabular}{lrrrrr|rrrrrr}
    \toprule
     &  \multicolumn{5}{c|}{\textbf{VerbMobil Gender}}   &  \multicolumn{5}{c}{\textbf{Yelp Gender }} \\
    \midrule
     & \textbf{Attr.} &\textbf{Utility}  & & & \textbf{} & \textbf{Attr.} &\textbf{Utility}  & & & \textbf{} \\
    \textbf{Method} & \textbf{F1$\downarrow$} &\textbf{F1$\uparrow$}  & \textbf{METEOR$\uparrow$} &\textbf{GAR$\uparrow$}  & $\bf{P_{Mean}}$$\uparrow$ & \textbf{F1$\downarrow$} &\textbf{F1$\uparrow$}  & \textbf{METEOR$\uparrow$} & \textbf{GAR$\uparrow$}& $\bf{P_{Mean}}$$\uparrow$\\
    \midrule
    %Original Test set & 71.03 & 59.73 & & 48.97  & 87.92 & 97.55 & & 29.11 \\
    Original Test set & 72.24 & 59.73 & 100 & 61.08 & 49.78 & 87.92 & 97.55 & 100 & 86.18 & 73.95 \\
    \midrule
    BT (DE) & 67.09 & \textbf{54.19}  & \textbf{41.21} & 68.16 &  49.12 & 82.37 & \textbf{95.45} & \textbf{52.42} & 88.83 & \textbf{63.58}\\
    BT (ES) & 62.58 & 50.47 & 31.38 &  77.01 & 49.07 & 76.51 & 91.54 & 38.89 & 90.37 & 61.07\\
    BT (FR) & 67.98 & 52.69 & 36.77 & 77.28 & 49.69 & 76.48 & 91.80 & 40.63 & 91.23 & 61.80 \\
    BT (JA) & 65.23 & 45.52 & 21.71 & 87.68 & 47.42 & 71.98 & 92.39 & 35.47 & 93.00 & 62.22 \\
    BT (RU) & 68.73 & 52.56 & 39.27 & 66.24 & 47.33 & 79.11 & 94.17 & 45.17 & 85.88 & 61.53 \\
    BT (ZH) & 63.96 & 51.70 & 25.80 & 91.79 & \textbf{51.33} & 72.85 & 92.22 & 34.40 & \textbf{95.57} & 62.34 \\
    \midrule
    CAE& \textbf{49.71} & 23.06 & 6.30 &  \textbf{93.70} & 43.32 & 68.72  & 88.37 & 40.18 & 60.38 & 55.05 \\
    BST& 66.51 & 18.06 & 1.69 & 23.18 & 19.11 & 51.00  & 71.00  & 24.03  & 58.45 & 50.62 \\
    UNMT & 59.97 & 29.49 & 13.95 & 45.99 & 32.37 & 68.92 & 90.71 & 45.67  & 77.65 & 61.28 \\
    DLS & 61.42 & 31.66 & 17.34 & 47.81 & 33.85 & 50.43 & 82.99 & 34.55 & 77.03 & 61.03 \\
    Tag\&Gen & 69.66 & 36.40 & 7.40 & 67.79 & 35.48 & \textbf{33.64} & 79.49 & 36.30 & 55.32 & 58.93 \\
    %Tag\&Gen & 69.35 & 36.40 & 30.19 & 32.18 & \textbf{33.64} & 79.49 & 36.30 & \textbf{53.01} \\
    %UPara~\cite{krishna-etal-2020-reformulating}++ & 56.36 & 39.97 & 29.55  \\
    
    \bottomrule
     \vspace{-4mm}
  \end{tabular}
  }
  \caption{Evaluation on VerbMobil (low-resource scenario) and Yelp. Comparing style transfer models and BT}
  \label{tab:verbmobil_result}
  \end{center}
\end{table*}
\subsection{Evaluation tasks and Metrics}
\label{sec:metric}
Style transfer models are usually evaluated on three tasks: Transfer style (or attribute) strength, content preservation, and fluency~\cite{jin2021deep}. Although, our desire is for the models to have a very good transfer attribute strength, other evaluation tasks are important since there are several downstream tasks that would benefit immensely from fluency and content preservation. For example, content preservation is critical for question answering systems, and fluency is desirable for dialog generation systems since we may not be able to generate fluent replies with non-fluent inputs. 

%To clarify why we need other metrics apart from privacy and utility preservation, there are other downstream tasks that need fluency and content preservation that we did not include in the paper e.g QA requires content preservation. Other tasks like dialog generation will also benefit immensely from fluency since we may not be able to generate fluent replies with non-fluent inputs. 

For \textbf{Transfer attribute strength}, we measure the success of the transfer by a drop in attribute F1-score (\texttt{Attr}) on the transformed test set. For \textbf{Content preservation}, we choose \texttt{METEOR} because it takes into account word stems, synonyms and paraphrase leading to better recall.   \textbf{Fluency} measures grammaticality. Following \citet{krishna-etal-2020-reformulating}, we compute grammaticality acceptance rate (\texttt{GAR}) using available fine-tuned models\footnote{\url{https://huggingface.co/textattack/roberta-base-CoLA}} trained on CoLA~\cite{warstadt-etal-2019-neural}. Lastly,  we introduce a new task, \textbf{Utility} (\texttt{Util}) to measure the performance of the transformed texts on an available downstream NLP task. Further details are in \autoref{sec:appendix_metric}. 
To measure the overall performance across all tasks, we compute an average of all the metrics ($\bf{P_{Mean}}$). For transfer attribute strength, we subtract attribute F1 from 100 i.e $(100-Attr)$ because the value is decreasing while others are increasing. 
We provide more details in \autoref{sec:appendix_metric}.

\subsection{Results}

\begin{table}[t]
 \footnotesize
  \vspace{-3mm}
 \begin{center}
 \scalebox{0.93}{
  \begin{tabular}{lrr}
    \toprule
    \textbf{Method}  & \textbf{Content preservation} & \textbf{Fluency} \\ 
    \midrule
    BT (JA) & 4.16 & 4.76 \\
    BT (ZH) & \textbf{4.19} & \textbf{4.78} \\
    DLS & 3.42 & 4.25 \\
    Tag\&Gen& 2.89 & 3.42 \\
   % YelpS (gender) & 2.6M & 373K & 5K & 500 & 4K \\
    
    \bottomrule
     \vspace{-5mm}
  \end{tabular}
  }
  \caption{Human evaluation of content preservation and fluency on BT (JA), BT (ZH), DLS, Tag\&Gen }
  \label{tab:hum_eval}
  \end{center}
\end{table}

\begin{table}[t]
 \footnotesize
  \vspace{-3mm}
  
 \begin{center}
 \scalebox{0.93}{
  \begin{tabular}{lp{60mm}}
    \toprule
    \textbf{Method}  & \textbf{sentence} \\ 
    \midrule
    Original & this hotel seems to be very poorly run. \\
\midrule
    BT (DE) & The hotel seems to be very poorly operated.
 \\
    BT (JA) & This hotel seems to be very poorly managed.
     \\
    BT (ZH) & This hotel looks terrible. \\

\midrule
    CAE &  this place is definitely very good.
\\
    BST &  this hotel seems poorly run.
\\
    UNMT &  this hotel seems to be very clean .\\
    DLS &  i was n't very impressed with this place.\\
    Tag\&Gen&  this hotel seems to be gorgeous run .
 \\

    \bottomrule
     \vspace{-7mm}
  \end{tabular}
  }
  \end{center}
  \caption{Sample sentences for BT (DE), BT (JA), BT (ZH), CAE, BST, DLS, UNMT, Tag\&Gen }
  \label{tab:sampl_sentence}
\end{table}

\paragraph{Automatic Evaluation}
%with race as sensitive attribute and sentiment classification as utility task
We compare the performance of the style transfer models and back-translation models in terms of attribute F1, utility F1, METEOR, and GAR on three datasets (DIAL, VerbMobil and Yelp). \autoref{tab:twitter_result} shows the performance on DIAL dataset. We observe a reduction of $7-22\%$ in attribute F1 by a simple back-translation, with Chinese (ZH) preserving more privacy while maintaining $95\%$ of the original utility and highest score ($81\%$) for fluency. German (DE) has better METEOR score and utility on average but sacrificed a lot of privacy. The BT (ZH) model has similar or better performance as the Adversarial training and SMDSP proposed by ~\cite{xu-etal-2019-privacy} in privacy preservation, utility and content preservation. However, we find style transfer methods have much better privacy preservation than BT models with $45-75\%$ reduction in attribute F1, but they sacrificed a lot in terms of utility, content preservation ($<30$ METEOR except Tag\&Gen) and fluency ($<45\%$ GAR), making them not practical for real-life applications. 

\autoref{tab:verbmobil_result} shows the result on VerbMobil dataset. The BT models leads to a reduction of $3.5-9.7\%$ in attribute F1 while maintaining over $86\%$ of the original utility F1. We also find them to achieve better performance in METEOR and GAR, although the models are applied in zero-shot settings. The style transfer models performed terribly since they typically require massive amounts of data \cite{li-etal-2019-domain} and might be skewed in a data-scarcity scenario (5k sentences for VerbMobil). One particular strength of our approach is that it requires no additional data and most suited for zero-shot settings. 

We also examined the performance of BT models on Yelp dataset. The style transfer models preserve more gender privacy  ($19-54\%$) than the BT models ($5-16\%$). However, they have much worse results in terms of utility and fluency. Overall, the $\bf{P_{Mean}}$ of BT models is often better than the style transfer models for all datasets. 

\paragraph{Human Evaluation}
We further performed human evaluation for the two best privacy-preserving BT models (ZH and JA) and style transfer models (DLS and Tag\&Gen) in terms of content preservation and fluency. We recruited three raters, who are volunteers from our research lab including authors of the paper to evaluate the models. The three volunteers rated 100 sentences per model i.e 400 sentences per rater. The volunteers were not paid for the rating, and were informed that they could in principle, choose to withdraw from the annotation without consequences. We provide the annotation guideline on Github\footnote{\url{https://github.com/uds-lsv/author-profiling-prevention-BT}}.

\autoref{tab:hum_eval} shows the average rating by three professional speakers of English language on 100 sentences in the Yelp dataset, we found out that ZH and JA are rated much higher in content preservation -- over $4$ (on a $1-5$ Likert scale) while maintaining near perfect fluency ($4.7$). The inter-agreement Krippendorff $\alpha$ of our human raters is $0.69$ for both content preservation and fluency. On the other hand, DLS and Tag\&Gen are rated lower on both evaluation tasks. Although, Tag\&Gen preserves privacy more on Yelp according to ~\autoref{tab:verbmobil_result}. %, it was rated much worse by human raters.  
\autoref{tab:sampl_sentence} shows an example sentence confirming the content preservation and fluency of our approach. We provide more examples in \autoref{sec:more_examples}.
%We provide an example sentence showing qualitatively the content preservation and fluency of our approach in \autoref{tab:sampl_sentence}.

\section{Conclusion}
In this paper, we propose a zero-shot way to effectively lower the risk of author
profiling through multilingual BT using off-the-shelf translation models. We compare our approach with different style transfer models, achieving the best overall performance using 
an automatic and a human evaluation while requiring no additional training data. In the future,  we will (1) analyze how the language choice and translation quality affects the privacy preservation in BT, (2) investigate more on other metrics that can be used to aggregate the the four evaluation metrics corresponding to transfer attribute strength, content preservation, fluency, and utility, and %, different from $P_{MEAN}$ to aggregate the four evaluation metrics corresponding to transfer attribute strength, content preservation, fluency, and utility, and 
(3) extend the zero-shot BT method with some supervision to improve privacy. %to make it more reliable.  %we will investigate the linguistic characteristics of the BT languages, and extend the transfer abilities of the BT model with suitable style transfer models while maintaining utility.

We highlight a few limitations of our work. First, back-translation transformation remove content style but does not necessarily replace attribute markers like style transfer models, for example, given a text ``me and my husband ...'', style transfer models are more likely to change ``husband'' to ``wife'' but back-translation will not. Second, our back-translation technique also inherit some of the problems of machine translation generated texts like hallucination~\cite{raunak-etal-2021-curious}. We provide examples highlighting these issues in \autoref{sec:more_examples}.

\section{Broader Impact Statement and Ethics}
This paper presents an approach to prevent author profiling of sensitive user attributes. We understand there are many ethical concerns around gender and race, however, our definition and evaluation of user traits are constrained by the available datasets we found in the literature. We did not collect any new data to show the strength of our approach. We hope our research helps to protect the profiling of under-represented groups and communities.

\section*{Acknowledgements}
The presented research has been funded by the European Union’s Horizon 2020 research and innovation programme project COMPRISE (\url{http://www.compriseh2020.eu/}) under grant agreement No. 3081705. We thank Dana Ruiter for providing the initial draft of the annotation guideline. We also thank members of the Spoken Language Systems Group and anonymous reviewers for their useful feedback on the paper.

\bibliography{anthology,custom}
\bibliographystyle{acl_natbib}

%\clearpage

\appendix
\section{Data Description}
\label{sec:appendix_data}
In this paper, we conduct experiments on three datasets (DIAL, VerbMobil, and Yelp) from Twitter social media, dialog conversations, and business reviews domains. Each of the datasets have either race or gender as the sensitive information, and sentiment classification or dialog act classification as the downstream NLP task to measure utility. \autoref{tab:datasets} shows the datasets and their splits: \textbf{Attribute Train}, training corpus for the attribute classifier; \textbf{Utility Train}, training corpus for an NLP task; 
\textbf{Style Train}, training corpus for style-transfer models, \textbf{Dev}, the development set, and the \textbf{Test} set. %The training data for all the splits are disjoint except for \textit{Style Train} in the Twitter and VerbMobil corpora.  

\paragraph{DIAL} created by \cite{blodgett-etal-2016-demographic} for dialectal tweets classification of African American (AAE) and Standard American English (SAE), and each tweet is assigned a predicted race information -- AA or White, and sentiment (pos/neg). We make use of the subset of the tweets~\cite{elazar-goldberg-2018-adversarial} with over 80\% confidence in race prediction. The final dataset has 180K tweets (90K each for AA and White race), 80K of the tweets are used for training the attribute classifier while the remaining 100K are used for training sentiment classifier and style transfer models. 
\paragraph{VerbMobil} corpus~\cite{weilhammer-etal-2002-multi} is a dialog corpus of human to human telephone conversation that are scheduling appointments. The English VerbMobil has over 10K utterances, with only 6,538 with gender information and 6,093 with dialog act (DA) information. We make use of 1,096 utterances with both gender information and DA as the test set, and others for training and \textbf{Dev}. We used the same training set for attribute classification and style transfer models due to limited data.    

\paragraph{Yelp} review corpus created by \cite{reddy-knight-2016-obfuscating} has gender annotation (male and female), we combined this dataset with another Yelp review corpus~\cite{Shen2017_cae} with only sentiment annotation. By automatically comparing the reviews in the two datasets, we created a \textbf{Dev} and \textbf{Test} set with 4K reviews each with both gender and sentiment information. This can be used for future research to evaluate the utility of Yelp Gender dataset.

\section{Evaluation tasks and Metrics}
\label{sec:appendix_metric}
Style transfer models are usually evaluated on three tasks: Transfer style (or attribute) strength, content preservation, and fluency~\cite{jin2021deep}.
\begin{enumerate}
    \item Transfer attribute strength (\texttt{Attr}): For a binary attribute, the goal is to generate a sentence of attribute 1 given an initial sentence with attribute 0. We measure the success of the transfer by a drop in attribute F1-score on the transformed test set. %Although, we cannot measure this task directly using back-translation. 
    \item Content preservation(\texttt{METEOR}): This is measured using automatic metrics like BLEU~\cite{papineni-etal-2002-bleu}, ROUGE~\cite{lin-2004-rouge}, and METEOR~\cite{banerjee-lavie-2005-meteor}. We choose METEOR because it has better correlation with human than BLEU that is commonly used. Also, it takes into account word stems, synonyms and paraphrase when computing the score leading to better recall. Recently, it has been popularly adopted by the style-transfer community.
    \item Fluency(\texttt{GAR}): measures grammaticality. In most cases, this is measured using perplexity on the transformed set. However, \citet{krishna-etal-2020-reformulating} proposed computing the grammaticality score from a classifier trained on Corpus of Linguistic Acceptability (CoLA)~\cite{warstadt-etal-2019-neural} instead of perplexity because it is unbounded and unnatural sentences with common words may have low perplexity. We compute grammaticality acceptance rate (GAR) using available fine-tuned models\footnote{\url{https://huggingface.co/textattack/roberta-base-CoLA}}.
    
    \item Utility(\texttt{Util}): we introduce a new task to measure the performance of the transformed texts on an available downstream NLP task. For example, DIAL dataset that is popularly used can also be evaluated for sentiment classification~\cite{xu-etal-2019-privacy}. Here, we also used the F1-score. 
\end{enumerate}

To measure the overall performance across all tasks, we compute an average of all the metrics (\textbf{$P_{Mean}$}), because all the metrics range from 0 to 100. For the transfer strength, we use (100-F1) since the value is decreasing. Specifically, we compute: 

\begin{small}
\begin{equation*}
    P_{Mean}= \dfrac{100-Attr + Util + METEOR + GAR}{4}
\end{equation*}
\end{small}

\section{More Examples:}
\label{sec:more_examples}
We provide more examples from the three datasets we considered: Yelp, VerbMobil and DIAL
\begin{table}[ht]
 \footnotesize
  \vspace{-3mm}
 \begin{center}
 \scalebox{0.99}{
  \begin{tabular}{ll}
    \toprule
    \textbf{Method}  & \textbf{sentence} \\ 
    \midrule
    Original & me and my husband love tokyo lobby ! \\
\midrule
    BT (DE) & me and my husband love Tokyo Lobby!
 \\
 BT (ES) & I and my husband love Tokyo Lobby!
 \\
 BT (FR) & I love the Tokyo lobby with my husband!
 \\
    BT (JA) & my husband and i love the Tokyo lobby!
 \\
 BT (RU) & I and my husband love the tokyo lobby!
 \\
    BT (ZH) & My husband and I love Tokyo's amusement park! \\

\midrule
    CAE &  me and my wife loves in san lobby !
\\
    BST &  my wife and i love the tokyo lobby.
\\
    UNMT &  me and my wife love the interior ! \\
    DLS &  it and my wife love this lobby ! \\
    Tag\&Gen&  me and my husband earned tokyo lobby !
 \\
    \bottomrule
  \end{tabular}
  }
  \end{center}
  \caption{Yelp: Sample sentences for BT and style transfer models }
  \label{tab:yelp_example}
\end{table}

\begin{table}[ht]
 \footnotesize
  \vspace{-3mm}
 \begin{center}
 \scalebox{0.99}{
  \begin{tabular}{ll}
    \toprule
    \textbf{Method}  & \textbf{sentence} \\ 
    \midrule
    Original & Lord , i hope this aint nobody i know ! \\
\midrule
    BT (DE) & Sir, I hope this is no one I know!
 \\
 BT (ES) & Lord, I hope that this is not anyone who knows!
 \\
 BT (FR) & Lord, I hope this is not someone I know!
 \\
    BT (JA) & God, I wish there was no one I knew!
 \\
 BT (RU) & God, I hope it's no one I know!
 \\
    BT (ZH) & Oh, my God, I wish this wasn't someone I knew! \\

\midrule
    CAE &  <unk> is so good
\\
    BST &  ENTITY hope not someone I know!
\\
    UNMT & Dear God , i hope this is not good ! I miss you \\
    DLS &  Lord I hope this would be pretty much ! ENTITY \\
    Tag\&Gen& Lord , i hope this is actually know
 \\
    \bottomrule
     \vspace{-2mm}
  \end{tabular}
  }
  \end{center}
  \caption{DIAL: Sample sentences for BT and style transfer models}
  \label{tab:dial_example}
\end{table}

\begin{table*}[t]
\footnotesize
  \vspace{-3mm}
 \begin{center}
 \scalebox{0.98}{
  \begin{tabular}{ll}
    \toprule
    \textbf{Method}  & \textbf{sentence} \\ 
    \midrule
    Original & that gives us plenty of time to chill out before the morning \\
\midrule
    BT (DE) & this gives us plenty of time to cool off before the morning
 \\
 BT (ES) & The Committee recommends that the State party ...
 \\
 BT (FR) & which gives us a lot of time to cool down before the morning
 \\
    BT (JA) & it gives us enough time to cool down in the morning.
 \\
 BT (RU) & that gives us plenty of time to rest until the morning
 \\
    BT (ZH) & So we can have a good rest before the morning \\

\midrule
    CAE &  that is good for me , I am going to be out of town , I am out of town
\\
    BST &  okay , that is fine , I am going to be out of town , I am out of town
    \\
    UNMT &  okay what do you say we meet on Monday , around two P M\\
    DLS &  okay what time what time will you have to go in Trier \\
    Tag\&Gen&  yeah that is fine for me how about the twenty seventh or twenty seventh or twenty seventh
 \\
    \bottomrule
     \vspace{-2mm}
  \end{tabular}
  }
  \end{center}
  \caption{VerbMobil: Sample sentences for BT and style transfer models }
  \label{tab:verbmobil_example}
\end{table*}

\end{document}